%% file: main_meta.tex
\documentclass[]{fairmeta}
\usepackage{xspace}

\makeatletter
\DeclareRobustCommand\onedot{\futurelet\@let@token\@onedot}
\def\@onedot{\ifx\@let@token.\else.\null\fi\xspace}

\makeatother

\usepackage{multirow}
\usepackage{amsfonts}
\usepackage[table,xcdraw]{xcolor}
\usepackage[normalem]{ulem}
\useunder{\uline}{\ul}{}

\newcommand{\tocite}[1]{\textcolor{red}{[TO CITE]}}

\newcommand{\modeln}{\textsc{Tuna-2}\xspace}
\newcommand{\modelp}{\textsc{Tuna-R}\xspace}
\newcommand{\tuna}{\textsc{Tuna}\xspace}
\newcommand{\model}{\textsc{Tuna-2}\xspace}

\title{\textcolor{metablue}{\model}: Pixel Embeddings Beat Vision Encoders for Multimodal Understanding and Generation}

\author[1,2,*]{\fontsize{9}{11}\selectfont Zhiheng Liu}
\author[1,3,*]{\fontsize{9}{11}\selectfont Weiming Ren}
\author[1]{\fontsize{9}{11}\selectfont Xiaoke Huang}
\author[1]{\fontsize{9}{11}\selectfont Shoufa Chen}
\author[1]{\fontsize{9}{11}\selectfont Tianhong Li}
\author[2]{\fontsize{9}{11}\selectfont Mengzhao Chen}
\author[2]{\fontsize{9}{11}\selectfont Yatai Ji}
\author[1]{\fontsize{9}{11}\selectfont Sen He}
\author[1]{\fontsize{9}{11}\selectfont Jonas Schult}
\author[1]{\fontsize{9}{11}\selectfont Belinda Zeng}
\author[1]{\fontsize{9}{11}\selectfont Tao Xiang}
\author[3]{\fontsize{9}{11}\selectfont Wenhu Chen}
\author[2]{\fontsize{9}{11}\selectfont Ping Luo}
\author[1]{\fontsize{9}{11}\selectfont Luke Zettlemoyer}
\author[1]{\fontsize{9}{11}\selectfont Yuren Cong}
\affiliation[1]{Meta AI}
\affiliation[2]{The University of Hong Kong}
\affiliation[3]{University of Waterloo}
\contribution[*]{Joint first authors, listed alphabetically by last name}

\abstract{
\input{sections/0.abs}
}

\date{April 28, 2026}

\metadata[Project page]{\url{https://tuna-ai.org/tuna-2}}
\vspace{8mm}

\begin{document}

\maketitle

\input{sections/1.intro_wm}
\input{sections/3.method.tex}
\input{sections/4.exp.tex}

\input{sections/2.related.tex}
\input{sections/5.conclusion.tex}

\clearpage
\newpage
\bibliographystyle{assets/plainnat}
\bibliography{ref}

\end{document}

%% file: sections/0.abs.tex
Unified multimodal models typically rely on pretrained vision encoders and use separate visual representations for understanding and generation, creating misalignment between the two tasks and preventing fully end-to-end optimization from raw pixels. We introduce \model, a native unified multimodal model that performs visual understanding and generation directly based on pixel embeddings. \model drastically simplifies the model architecture by employing simple patch embedding layers to encode visual input, completely discarding the modular vision encoder designs such as the VAE or the representation encoder. Experiments show that \model achieves state-of-the-art performance in multimodal benchmarks, demonstrating that unified pixel-space modelling can fully compete with latent-space approaches for high-quality image generation. Moreover, while the encoder-based variant converges faster in early pretraining, \model's encoder-free design achieves stronger multimodal understanding at scale, particularly on tasks requiring fine-grained visual perception. These results show that pretrained vision encoders are not necessary for multimodal modelling, and end-to-end pixel-space learning offers a scalable path toward stronger visual representations for both generation and perception.

%% file: sections/1.intro_wm.tex
\begin{figure*}[h!]
\vspace{0.5em}
\centering
\includegraphics[width=1.00\linewidth]{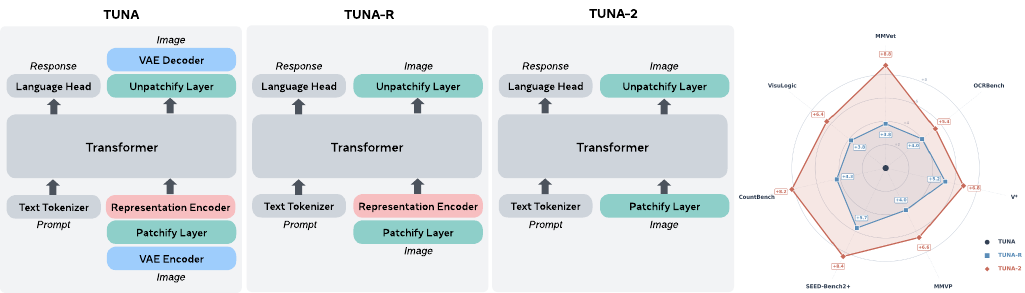}
\vspace{-2em}
\caption{Evolution of \modeln architecture and multimodal performance comparison. We simplify \tuna~\citep{liu2025tuna} by progressively stripping away its visual encoding components. By removing the VAE, we first derive \modelp, a pixel-space UMM that relies solely on a representation encoder. \modeln further streamlines the design by bypassing the representation encoder entirely, utilizing direct patch embedding layers for raw image inputs. \modeln using pixel embeddings outperforms both \modelp and \tuna across a diverse suite of multimodal benchmarks.}
\label{fig:framework}
\end{figure*}

\section{Introduction}\label{sec:intro}

Visual understanding and generation are two core capabilities in multimodal AI. Recent work has increasingly focused on native unified multimodal models (UMMs)~\citep{zhou2024transfusion, deng2025emerging, liu2025tuna}, which aim to integrate both capabilities within a single framework. A central challenge in building such models is encoding input images into visual representations that effectively support both understanding and generation. Early approaches~\citep{deng2025emerging, chen2025janus} adopted decoupled representations, using representation encoders such as CLIP~\citep{radford2021learning} for understanding and reconstruction-oriented encoders such as VQ-VAE~\citep{esser2021taming} for generation. To address the representation mismatch introduced by this design, more recent UMMs~\citep{xie2025show, liu2025tuna} have moved toward modelling both tasks using unified visual representations through a shared vision encoder.

Despite the significant progress, both decoupled and unified visual representation designs still rely heavily on pretrained vision encoders \citep{wan2025wan, tschannen2025siglip} for visual feature extraction. In parallel, recent research on multimodal understanding and generation has begun to move away from encoder-based modular designs toward simpler monolithic, encoder-free architectures. In multimodal understanding, newer native vision-language models \citep{diao2025pixels} remove the pretrained representation encoder and instead align images and natural language within a unified, end-to-end architecture. In visual generation, pixel-space diffusion models~\citep{hoogeboom2023simple, chen2025pixelflow, li2025back} have shown increasing flexibility, stronger scalability, and state-of-the-art performance on a wide range of tasks, suggesting that pretrained VAE encoders may no longer be essential even for high-fidelity image synthesis.


Motivated by these observations, we ask a natural but largely unexplored question: can we move beyond pretrained vision encoders altogether, and build unified multimodal models through end-to-end native modelling directly from raw pixels?

We answer this question affirmatively by introducing \model, a native unified multimodal model that attempts to progressively simplify the encoder modules, and ultimately remove vision encoders completely. We first introduce \modelp, which eliminates the VAE model while keeping a representation encoder in the model architecture. \modelp performs multimodal understanding similar to standard encoder-based LMMs, and supports visual generation through pixel-space flow matching with an $x$-prediction objective. We then propose \modeln, which further simplifies the architecture by removing the encoder entirely and using only a single transformer decoder to process image and video tokens. As a result, \modeln enables end-to-end native unified modelling directly from raw pixels, without relying on any pretrained encoder modules.

\begin{figure}[]
\vspace{-2em}
\centering
\includegraphics[width=0.9\linewidth]{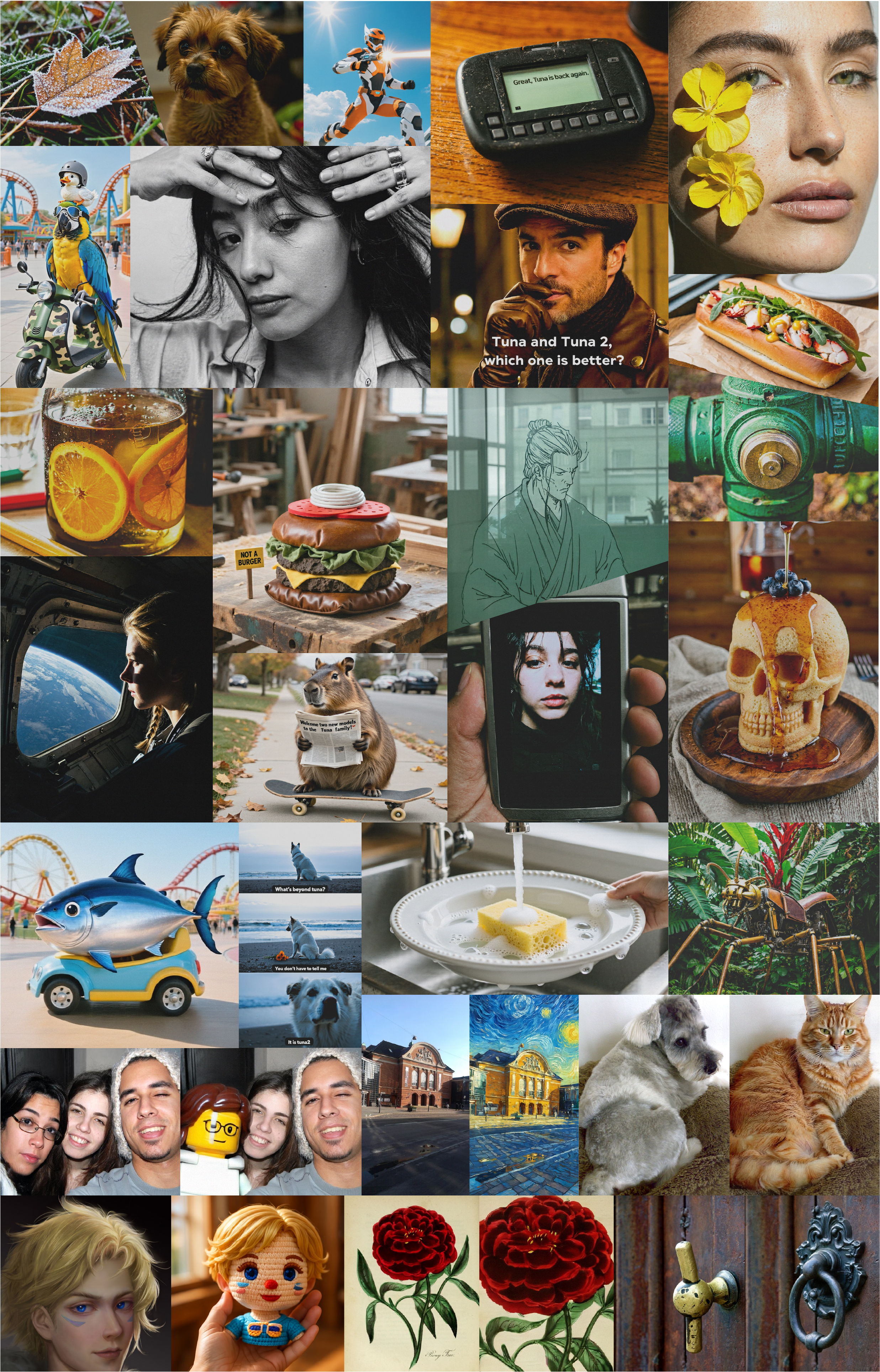}
\vspace{-0.8em}
\caption{While being completely encoder-free, \model is capable of performing high-fidelity text-to-image generation and image editing.}
\label{fig:teaser}
\end{figure}

Since learning unified representations directly in high-dimensional pixel space is substantially more challenging than learning them in a compact latent space, we further introduce a masking-based visual feature learning scheme to stabilize training and encourage the learning of more robust pixel-space representations. Together, these designs enable \model to achieve state-of-the-art performance across a diverse set of multimodal understanding and generation benchmarks. More importantly, our controlled comparison reveals a clear design insight: after sufficient visual pretraining, the encoder-free \modeln becomes competitive with the encoder-based \modelp on visual generation, while consistently outperforming it on multimodal understanding, especially on benchmarks that require fine-grained visual perception. These findings suggest that removing pretrained vision encoders can be advantageous for learning stronger fine-grained visual representations in end-to-end pretraining. As shown in Figures~\ref{fig:framework} and~\ref{fig:teaser}, this leads to highly competitive performances in both multimodal understanding and generation.

Our main contributions are summarized as follows:
\begin{itemize}
 \item [1.] We propose \model, a native unified multimodal model that supports multimodal understanding and generation with encoder-free designs, achieving state-of-the-art performance across a wide range of understanding and generation benchmarks.
 
\item [2.] We conduct controlled comparisons between \model and an encoder-based pixel-space UMM variant \modelp, and show that after sufficient multimodal pretraining, \model and its encoder-free design are competitive on generation and advantageous for understanding, especially on fine-grained, perception-intensive tasks.

\item [3.] We conduct comprehensive ablations and analyses on pixel-space UMMs to study their training dynamics and model behaviours, offering useful insights for the development of future native unified multimodal models.
\end{itemize}

\clearpage

\clearpage

%% file: sections/3.method.tex
\section{Method}\label{sec:method}

In this section, we present \model, a native unified multimodal model that performs visual understanding and generation both in pixel space. We start by detailing our approach to progressively remove vision encoder components to derive \model in Section~\ref{sec:arch}. We then describe our masked feature learning scheme in Section~\ref{sec:mask} and our model training pipeline in Section~\ref{sec:training}.

\subsection{Towards Encoder-Free Unified Models}\label{sec:arch}
As shown in Figure~\ref{fig:framework}, existing UMMs with unified visual representations, such as \tuna \citep{liu2025tuna}, typically consist of a vision encoder and an LLM decoder for joint vision-language modeling, followed by modality-specific heads, including a language modelling head for autoregressive text generation and a flow matching head for image generation. In this work, we propose \model as an encoder-free UMM formulation by progressively simplifying the vision encoder components in existing architectures. Our design process for this architectural simplification is as follows:

\noindent\textbf{Representation encoder-based architecture.} First, we attempt to remove the VAE model and only employ a pretrained representation encoder in the vision encoder. As shown in Figure~\ref{fig:framework}, this resonates a standard paradigm for vision-language modelling: the representation encoder first encodes input images into visual tokens, which are then combined with the text tokens in the LLM decoder for joint vision-language modelling. Originally proposed in LLaVA \citep{liu2023visual}, this paradigm has been verified and scaled up by later works such as Qwen3-VL \citep{bai2025qwen3} and InternVL3.5 \citep{wang2025internvl3}, and remains the most popular framework for multimodal understanding. We refer to this intermediate design as \textbf{\modelp}. Although our ultimate goal is to move beyond encoder-based architectures, we view \modelp as an important intermediate step that enables a controlled comparison with \modeln.

\noindent\textbf{Encoder-free (non-encoder) architecture.} Second, we consider a further simplified architecture that removes the representation encoder entirely, which becomes our main design for \textbf{\modeln}. As shown in Figure~\ref{fig:framework}, this design replaces pretrained vision encoders with simple patch embedding layers that convert images into visual tokens, which are then processed jointly with text tokens by the LLM decoder. Similar encoder-free designs have recently been explored in models such as Mono-InternVL \citep{luo2025mono} and NEO \citep{diao2025pixels}. By removing the pretrained representation encoder, this design avoids its built-in inductive biases, such as fixed input resolutions and limited access to fine-grained low-level visual details. It also simplifies the model architecture into a single unified transformer. In Section~\ref{sec:exp}, we present a series of in-depth analyses comparing \modeln with \modelp, and demonstrate the effectiveness and scalability of \modeln.

\noindent\textbf{Pixel-space image generation.} Our VAE-free design allows us to directly perform multimodal understanding and text generation using the LLM decoder and the language modelling head. However, discarding the VAE also means that we can no longer adopt the designs from existing UMMs and generation-only models that follow the latent diffusion architecture. To effectively perform pixel-space image generation, we adopt the $x$-prediction and $v$-loss paradigm from JiT \citep{li2025back} for pixel-space flow matching. Specifically, given the source image $x_1$, the sampled noise $x_0\sim \mathcal{N}(\mathbf{0}, \mathbf{I})$ and the timestamp $t$, we employ rectified flow \citep{liu2022flow, lipman2022flow} and its linear schedule to construct a noisy sample in pixel space:
\begin{equation}
    x_t = tx_1 + (1-t)x_0, t\in[0, 1].
\end{equation}
\model is then formulated to directly predict the clean image from the noisy image in pixel space:
\begin{equation}
    x_\theta = \pi_\theta(x_t, c, t),
\end{equation}
where $\pi_\theta$ is our unified model (vision-language backbone and flow matching head) and $c$ is the conditioning signals (text for text-to-image generation and text+image for image editing). As suggested in JiT, while our model directly predicts $x_\theta$, we still transform it into the velocity term $v_\theta$ and regress $v_\theta$ as our learning objective:
\begin{align}
    v_\theta &= \frac{x_\theta-x_t}{1-t}, \label{eq:velocity}\\
    \mathcal{L}_{\mathrm{flow}}&=\mathbb{E}_{t, c, x_1, x_0}||v_\theta - v||^2_2,
\end{align}
where $v$ is the ground truth velocity defined by $v=x_1-x_0$. During inference, we employ the Euler solver and predict the denoised image at $t^\prime$ from the noisier image at $t<t^\prime$ based on the velocity term $v_\theta$, such that $x_{t^\prime} = x_{t} + (t^\prime - t)v_\theta$, where $v_\theta$ is transformed from our model prediction $x_\theta$, based on Equation~\ref{eq:velocity}.

\begin{figure*}[t!]
\centering
\includegraphics[width=1.00\linewidth]{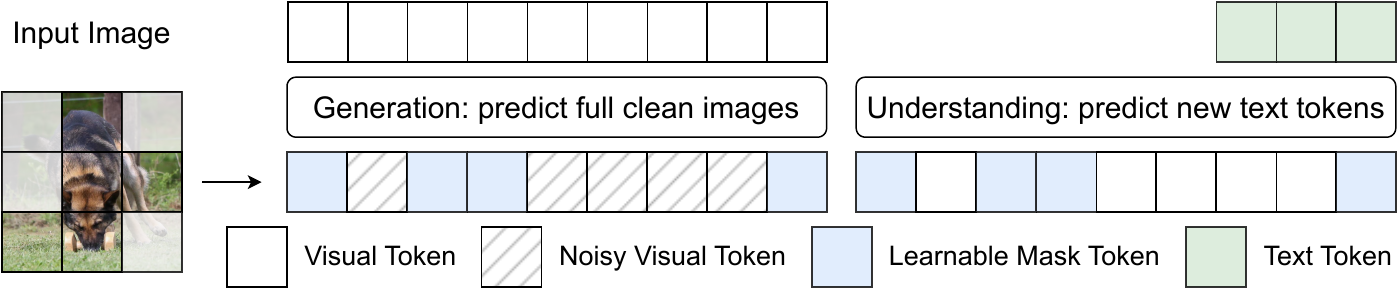}
\vspace{-2em}
\caption{Illustration of our proposed masking-based feature learning scheme. During training, we use the learnable mask token to regularize multimodal understanding and perform masked prediction for visual generation.}
\label{fig:mask}
\vspace{-1em}
\end{figure*}

\subsection{Learning Robust Visual Representations via Masking}\label{sec:mask}
While removing the VAE simplifies our model architecture and enables fully end-to-end unified multimodal training, it also shifts visual modelling from a compact latent space to the much higher-dimensional pixel space. As a result, learning a unified visual representation becomes more challenging: the increased redundancy in pixel-space inputs makes it easier for the model to rely on superficial shortcuts, rather than learning visual cues that are genuinely informative for both understanding and generation. To learn more robust visual representations in pixel space, we introduce a masking-based visual feature learning scheme. As shown in Figure~\ref{fig:mask}, during training, we (optionally) randomly select a subset of image patches according to a masking ratio and replace the masked visual tokens with a learnable mask token before feeding them into the LLM decoder. The same masking operation is applied to both generation and understanding examples, but plays different roles in the two settings:
\begin{itemize}
    \item [1.] For generation examples, we let the model predict the clean image patches in \textit{both the masked and the unmasked regions}, such that (1) we create a harder denoising problem for the model to predict clean images from partially observed noisy images; and (2) it encourages the learnable mask token to absorb useful information for image reconstruction conditioned on the visible context.
    \item [2.] For understanding examples, our model predicts the ground truth text response based on the \textit{masked visual input}. In this case, masking serves as a regularization mechanism that forces the model to perform multimodal reasoning under partial visual observation, leading to more robust visual representations.
\end{itemize}
Our masking-based feature learning scheme resembles masked modelling methods in visual understanding and generation, such as MAE \citep{he2022masked} and SigLIP 2 \citep{tschannen2025siglip} for semantic learning and MaskGIT \citep{chang2022maskgit} and DeTok \citep{yang2025latent} for visual generation.
Empirically, we find that applying masking leads to enhanced model performance during pretraining stages.

\subsection{Training Pipeline}\label{sec:training}
Our encoder-free design enables fully end-to-end training of \model, without requiring separate stages to train connector layers, which is a common design in encoder-based modular approaches. As described below, our training pipeline consists of two stages, both of which are carried out in a fully end-to-end manner:

\noindent\textbf{Stage 1: full model pretraining.} In the first stage, we aim to establish a strong initialization for the flow matching head, and adapt pixel-space visual inputs for unified multimodal understanding and generation. To achieve this, we train the full model jointly on two tasks: image captioning and text-to-image generation.

\noindent\textbf{Stage 2: supervised finetuning (SFT).} Next, we perform supervised fine-tuning (SFT) of the full model with a lower learning rate. We use datasets for image editing, image instruction-following, and high-quality image generation. This step refines \model's abilities, boosting performance and generalization across various multimodal tasks.

For \modelp, which includes a connector layer between the representation encoder and the LLM decoder, we add an extra alignment stage before Stage 1. In this stage, we train only the connector layer for a short period using image captioning and text-to-image generation data. As noted above, \modeln does not require this additional stage because of its encoder-free design.





%% file: sections/4.exp.tex
\section{Experiments}\label{sec:exp}


\subsection{Experiment Setup}
We employ Qwen2.5-7B-Instruct~\citep{qwen2024qwen2} as the LLM decoder and use a patch embedding size of 16 for \modeln. For Stage 1 pretraining, we use 550M in-house image-text pairs, consisting of 30\% image captioning data for multimodal understanding and 70\% text-to-image generation data. In addition, we include text-only data from Nemotron \citep{bercovich2025llama}, which accounts for 20\% of the total pretraining data. The full model is trained end-to-end for 300k steps on 64 nodes with the AdamW optimizer~\citep{loshchilov2017decoupled} and a learning rate of $1 \times 10^{-4}$. We employ the proposed masking-based feature learning strategy during the final 40\% of pretraining, applying masking in 50\% of training examples and randomly sampling the masking ratio from 0\% to 50\%. For Stage 2 supervised finetuning, we use a curated SFT corpus covering image instruction-following, image editing, and high-quality image generation. Specifically, for image instruction-following, we include 13M conversational examples from the open-source FineVision~\citep{wiedmann2025finevision} dataset. For image editing, we use approximately 2M examples from OmniEdit~\citep{wei2024omniedit}. This stage is trained for 50k steps with AdamW and a learning rate of $2 \times 10^{-5}$. For all training stages, we pad the input sequence length to 16k tokens per GPU.

For \modelp, we use the same Qwen2.5-7B-Instruct as the LLM decoder. We follow \tuna and adopt SigLIP 2 So400M~\citep{tschannen2025siglip} as the representation encoder. For the connector-alignment stage in \modelp, we train the model for 3k steps with AdamW and a learning rate of $5 \times 10^{-4}$.

\input{tables/table_image_und}
\input{tables/table_gen}

\input{tables/table_image_third_party}

\subsection{Main Results}
\label{sec:main_results}

\noindent\textbf{Image understanding.} We employ a comprehensive evaluation suite consisting of nine VQA benchmarks, including GQA \citep{hudson2019gqa}, RealWorldQA \citep{xai2024-grok15v}, MMVet \citep{yu2023mm}, MMMU \citep{yue2024mmmu}, MMVP \citep{tong2024eyes},  SEED-Bench2\texttt{+} \citep{li2024seed}, AI2D \citep{kembhavi2016diagram}, ChartQA \citep{masry2022chartqa} and OCRBench \citep{liu2024ocrbench}, to evaluate the image understanding capabilities for \model. As shown in Table~\ref{tab:img_und}, after removing the VAE model, both \modelp and \modeln outperform \tuna and achieve state-of-the-art results among all 7B-scale native UMMs, demonstrating the effectiveness of our pixel-space unified representations. Notably, \modeln outperforms \modelp even after replacing the representation encoder with the simple patchify layer. This may indicate that large-scale joint training of a unified, monolithic architecture achieves better understanding performance than relying on the inductive bias in the pretrained representation encoders in the modular setting. We show more analysis comparing \modelp and \modeln in Section~\ref{sec:analysis_arch}.

To further understand the benefit of our pixel-space unified models, we include several pixel-centric benchmarks that focus heavily on visual reasoning over fine-grained visual details (e.g. recognizing very small objects in high-resolution images). These benchmarks include V* \citep{wu2024v}, CountBench \citep{paiss2023teaching} and VisuLogic \citep{xu2025visulogic}. As shown in Table~\ref{tab:img_und}, both \modelp and \modeln outperform latent-space UMMs (e.g. Show-o2, \tuna, etc.) across all benchmarks, indicating the necessity of pixel-space visual representations when reasoning over fine-grained visual details.

\noindent\textbf{Image generation.} We evaluate the image generation performance of \model on GenEval~\citep{ghosh2023geneval} and DPG-Bench~\citep{hu2024ella}. As shown in Table~\ref{tab:geneval_dpg}, both \modelp and \modeln achieve state-of-the-art results on these benchmarks and perform competitively with contemporary approaches such as BAGEL, Mogao, and \tuna. Between the two variants, \modelp consistently performs slightly better than \modeln, suggesting that the semantic prior introduced by the representation encoder helps our method learn a stronger image generation model. Overall, these results show that even without a VAE and while performing image generation entirely in pixel space, \model remains competitive with recent state-of-the-art unified models, highlighting the effectiveness and scalability of our pixel-space generation design.

Since existing image generation benchmarks such as GenEval mainly evaluate text-image alignment and the model's world knowledge, we further adopt an LLM-judge-based evaluation to assess the quality and diversity of images generated by \modeln, \modelp, and \tuna. Specifically, we sample 1.5K text prompts and ask each model to generate four images per prompt. We then use GPT-5.4 \citep{openai2026introducinggpt54} and Claude Opus 4.7 \citep{anthropic2026introducingclaudeopus47} as judges to select the best model among the three for each prompt based on the quality and diversity of the generated images, where quality mainly refers to realism and fine-grained detail and texture fidelity, while diversity measures the extent to which the four generated images exhibit distinct visual variations under the same prompt. The results are reported in Table~\ref{tab:llm_eval}. We find that under both LLM judges, \modeln achieves competitive image generation quality, performing comparably to \modelp and better than \tuna, while being significantly preferred in terms of diversity. These results demonstrate that our encoder-free design is highly effective and enables \modeln to generate both high quality and diverse images.

\begin{figure*}[t!]
\centering
\begin{minipage}[c]{0.40\linewidth}
\centering
\vspace{0.1em}
\captionof{table}{Image reconstruction performance for different visual tokenizers.}

\label{tab:tokenizer_comparison}
\vspace{-1em}
\setlength{\tabcolsep}{3pt}
\renewcommand{\arraystretch}{1.05}
\resizebox{\linewidth}{!}{%
\begin{tabular}{lcccc}
\toprule
\textbf{Tokenizer} & \textbf{Res.} & \textbf{rFID$\downarrow$} & \textbf{PSNR$\uparrow$} & \textbf{SSIM$\uparrow$} \\
\midrule
\multicolumn{5}{l}{\textit{Specialized tokenizers}} \\
SD-VAE              & 256 & 1.06 & 28.62 & 0.86 \\
GigaTok             & 256 & 0.51 & 21.32 & 0.69 \\
VA-VAE              & 256 & 0.26 & 28.59 & 0.80 \\
DC-AE               & 512 & 0.22 & 26.15 & 0.71 \\
MAE-Tok             & 512 & 0.62 & --    & --   \\
TexTok              & 512 & 0.73 & 24.45 & 0.66 \\
FLUX.1[dev]-VAE$^{\dagger}$ & 512 & 0.06 & 33.65 & 0.93 \\
\midrule
\multicolumn{5}{l}{\textit{Unified tokenizers}} \\
UniTok              & 256 & 0.38 & --    & --   \\
TokenFlow           & 384 & 0.63 & 22.77 & 0.73 \\
X-Omni$^{\dagger}$  & 512 & 8.30 & 15.66 & 0.38 \\
MingTok$^{\dagger}$ & 512 & 0.53 & 23.49 & 0.61 \\
RAE                 & 256 & 0.61 & 19.20 & 0.44 \\
PS-VAE              & 256 & 0.20 & 28.79 & 0.82 \\
\textbf{\modelp}    & 512 & \textbf{0.12} & 32.22 & \textbf{0.93} \\
\textbf{\modeln}    & 512 & 0.15 & \textbf{32.80} & \textbf{0.93} \\
\bottomrule
\end{tabular}%
}
\end{minipage}%
\hfill
\begin{minipage}[c]{0.58\linewidth}
\centering
\includegraphics[width=\linewidth]{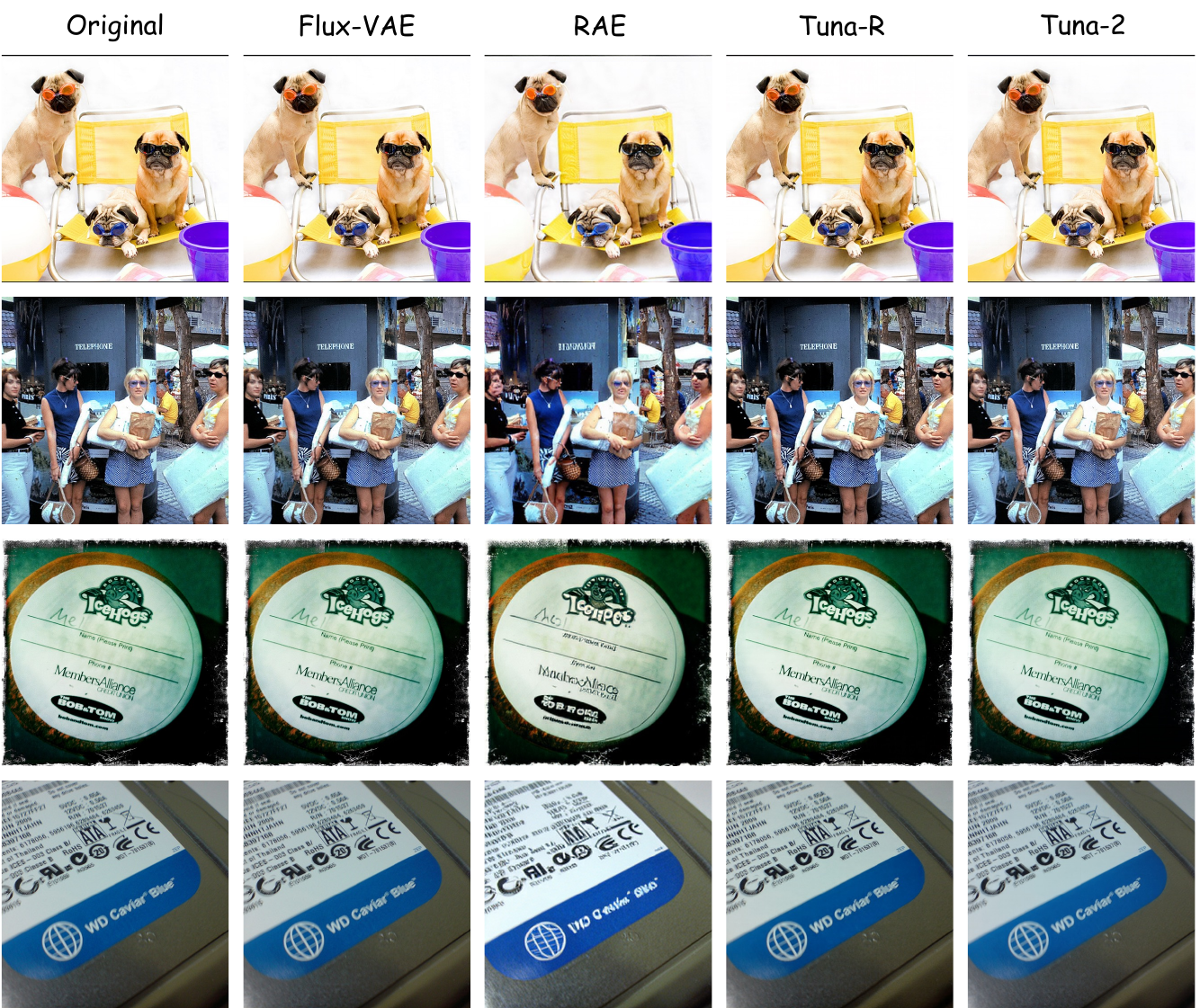}
\vspace{-2em}
\captionof{figure}{Qualitative results for different visual tokenizers.}
\label{fig:recon}
\end{minipage}
\vspace{-1em}
\end{figure*}

\noindent\textbf{Image editing.} 
We further evaluate the image editing capability of \model on ImgEdit \citep{ye2025imgedit}. As shown in Table~\ref{tab:img_edit}, \modeln achieves strong editing performance among unified models, outperforming earlier baselines such as OmniGen \citep{xiao2025omnigen}, BAGEL \citep{deng2025emerging}, UniWorld \citep{lin2025uniworld}, and OmniGen2 \citep{wu2025omnigen2}. While being slightly behind \tuna and \modelp, \modeln remains competitive with strong generation-only and unified editing systems, despite performing editing directly in pixel space without relying on vision encoders. These results suggest that our pixel-space unified modelling framework can effectively support instruction-guided image editing, while the small gap between \modeln and encoder-based variants indicates that pretrained visual priors may still provide benefits for fine-grained editing fidelity.

\noindent\textbf{Image reconstruction.} To further investigate the image generation capability of \model, we examine whether the model can faithfully reconstruct an input image from its corresponding pixel-space visual representations. To this end, we perform lightweight finetuning on an image reconstruction task and evaluate reconstruction quality on the ImageNet validation set~\citep{deng2009imagenet}. As shown in Table~\ref{tab:tokenizer_comparison}, both \modelp and \modeln achieve strong reconstruction performance, ranking first among unified tokenizers and approaching specialized image tokenizers such as the VAE model in FLUX.1 [dev] \citep{batifol2025flux}. Figure~\ref{fig:recon} further shows that our models produce substantially better reconstructions than other non-KL-regularized VAE approaches such as RAE \citep{zheng2025diffusion}. These results indicate that our pixel-space unified visual representation can support strong image reconstruction and generation quality even without relying on VAE models.



\begin{figure*}[t!]
\centering
\includegraphics[width=1.00\linewidth]{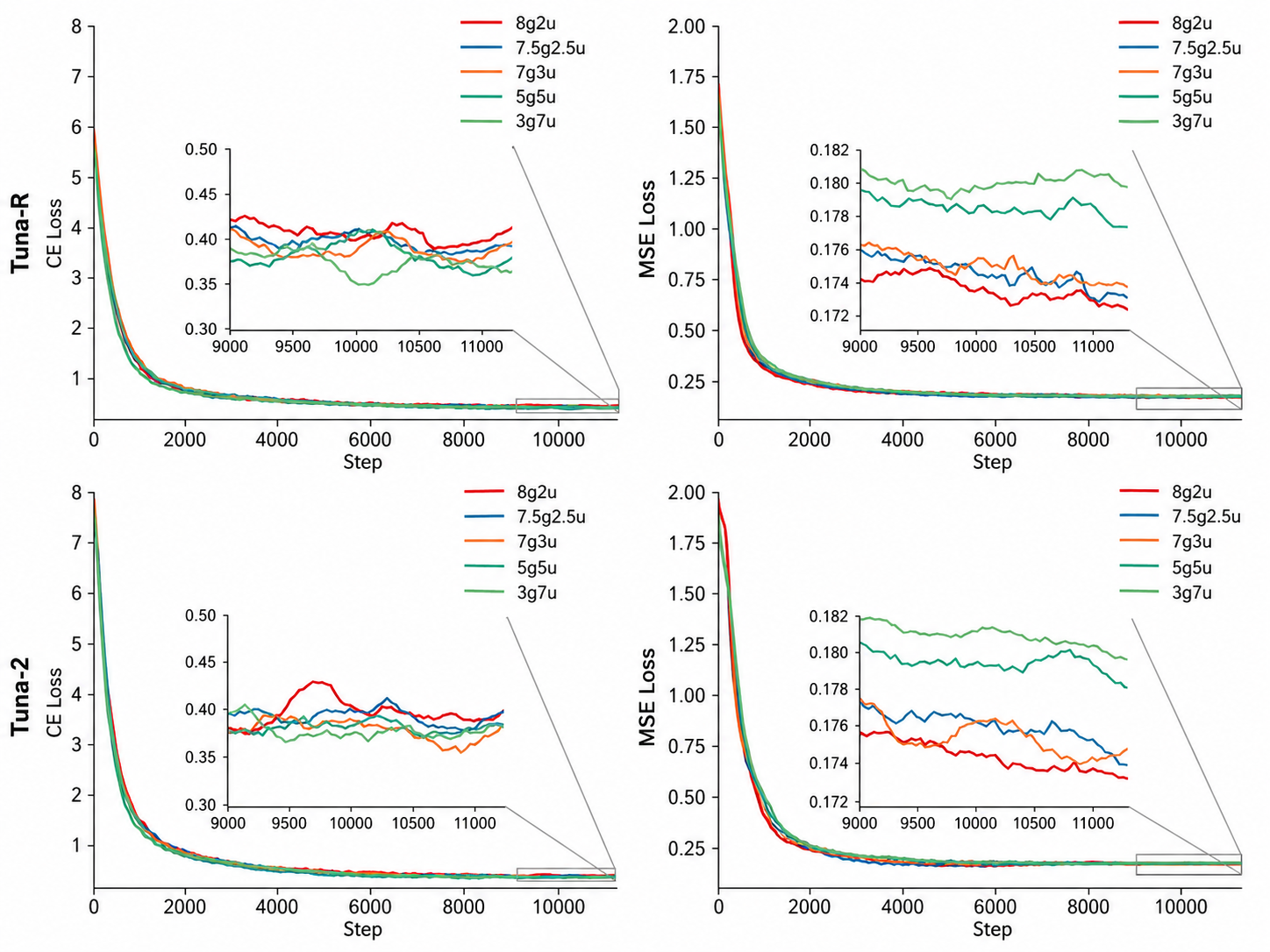}
\vspace{-2.5em}
\caption{Comparison of the image generation (MSE) and language modelling (CE) loss curves based on different understanding-generation data ratios during model training.}
\label{fig:ratio}
\end{figure*}

\subsection{Ablation: Model Training Dynamics}
\label{sec:analysis_training}
To better understand the learning dynamics of our joint multimodal training paradigm, we conduct a series of pretraining ablation studies on both \modelp and \modeln by varying the sampling ratio between generation and understanding data. We use the notation \texttt{xgyu} (e.g., \texttt{8g2u}) to represent a generation-to-understanding sampling ratio of $x:y$. As shown in Figure~\ref{fig:ratio}, increasing the proportion of either generation or understanding data consistently reduces its corresponding training loss, namely MSE for flow matching and cross entropy loss (CE) for language modelling. Notably, the MSE loss is more sensitive to changes in the sampling ratio, while the CE loss varies within a relatively smaller range across different ratios. This suggests that both tasks benefit from scaling up the amount of their corresponding training data, with the generation objective being more affected by the data mixture. We also observe that a generation-to-understanding ratio of $7:3$ (\texttt{7g3u}) achieves the best trade-off between the two objectives, yielding a strong balance between generation and understanding performance. We therefore adopt this data sampling ratio in all experiments.

\input{tables/table_mask}

\subsection{Ablation: Masking-based Feature Learning}
To verify the effectiveness of our proposed masking-based feature learning strategy, we conduct controlled experiments on both \modelp and \modeln using the smaller Qwen-2.5-Instruct-1.5B backbone. Since we expect this strategy to serve as a representation enhancement strategy after the model has acquired basic multimodal knowledge, we do not apply this objective from the beginning of pretraining. Instead, we first train all models for 50k steps; We then split each model variant into two groups: one continues standard pretraining, while the other continues pretraining with the masking-based feature learning objective applied with a probability of 50\%. Both groups are further trained for another 50k steps.
As shown in Table~\ref{tab:ablation_mask}, this strategy consistently improves performance on both understanding and generation benchmarks for both model variants. We further observe that \modeln benefits more from masked training than \modelp. We hypothesize that this difference is related to the SigLIP 2 representation encoder used in \modelp, since SigLIP 2 itself is pretrained with a similar masked prediction objective. In addition, the results suggest that compressing images into the VAE latent space before visual encoding (\tuna) can introduce certain information loss for visual understanding, compared with directly encoding pixel-level inputs through the vision encoder (\modelp).
Based on these findings, we apply the masking-based feature learning strategy during the final 40\% of pretraining, encouraging more robust visual representation learning for both multimodal understanding and generation.

\begin{figure*}[h!]
\vspace{1em}
\centering
\includegraphics[width=1.00\linewidth]{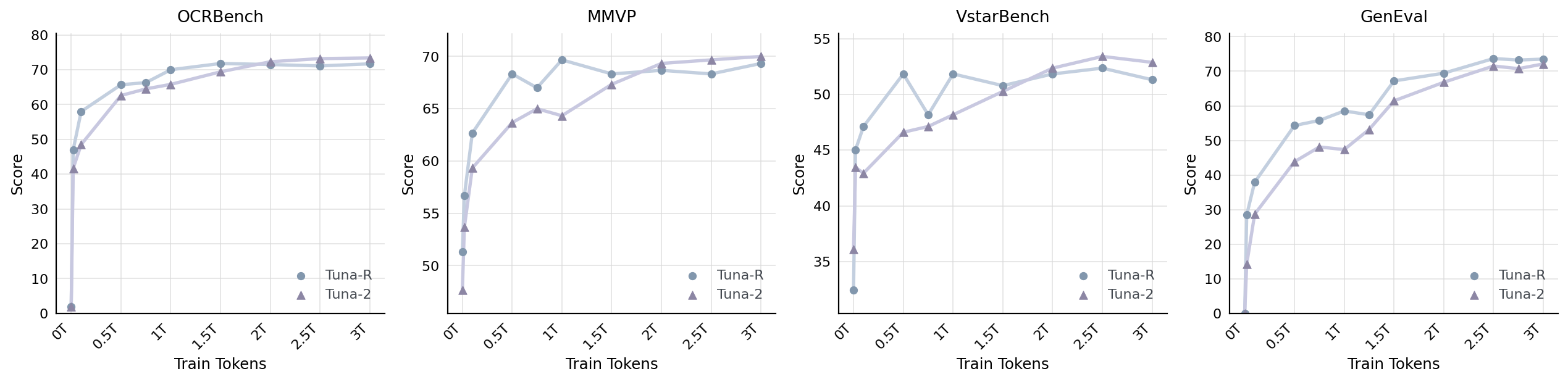}
\vspace{-2em}
\caption{Model accuracy curves with respect to training dataset size (measured by the number of tokens consumed by the model) for \modelp and \modeln.}
\label{fig:acc_curve}
\vspace{-1em}
\end{figure*}

\subsection{Analysis: \modelp vs. \modeln}
\label{sec:analysis_arch}
Our experimental results in Section~\ref{sec:main_results} reveal two interesting findings: First, without vision encoders, \modeln outperforms \modelp on most image understanding benchmarks. Second, \modelp achieves similar or slightly stronger performance than \modeln on generation tasks. To better understand these trends, we plot the performance curves of both models on understanding and generation benchmarks as the training data scale increases. As shown in Figure~\ref{fig:acc_curve}, on the three understanding benchmarks (OCRBench \citep{liu2024ocrbench}, MMVP \citep{tong2024eyes} and V* \citep{wu2024v}), \modelp actually outperforms \modeln during the early stage of the training. We believe this advantage comes from the pretrained representation encoder in \modelp, whose rich semantic priors help the model acquire multimodal understanding capabilities more quickly at the beginning of training. Nevertheless, \modeln later catches up and eventually surpasses \modelp. This suggests that the monolithic, encoder-free design of \modeln may be better suited to benefit from large-scale unified multimodal pretraining, enabling it to develop stronger multimodal understanding capabilities.

For generation evaluation on GenEval \citep{ghosh2023geneval}, we observe that \modelp consistently outperforms \modeln throughout the entire training process. This suggests that the semantic priors provided by the representation encoder play an important role in improving generation performance, which is consistent with prior findings from REPA \citep{yu2024representation} and \tuna \citep{liu2025tuna}. However, this trend gradually weakens as the training data scale increases. As shown in the figure, the two model variants achieve nearly identical performance after SFT. Overall, compared with latent-space generation, both \modelp and \modeln's pixel-space generation paradigm achieves competitive performances.

\begin{figure*}[t!]
\centering
\includegraphics[width=1.00\linewidth]{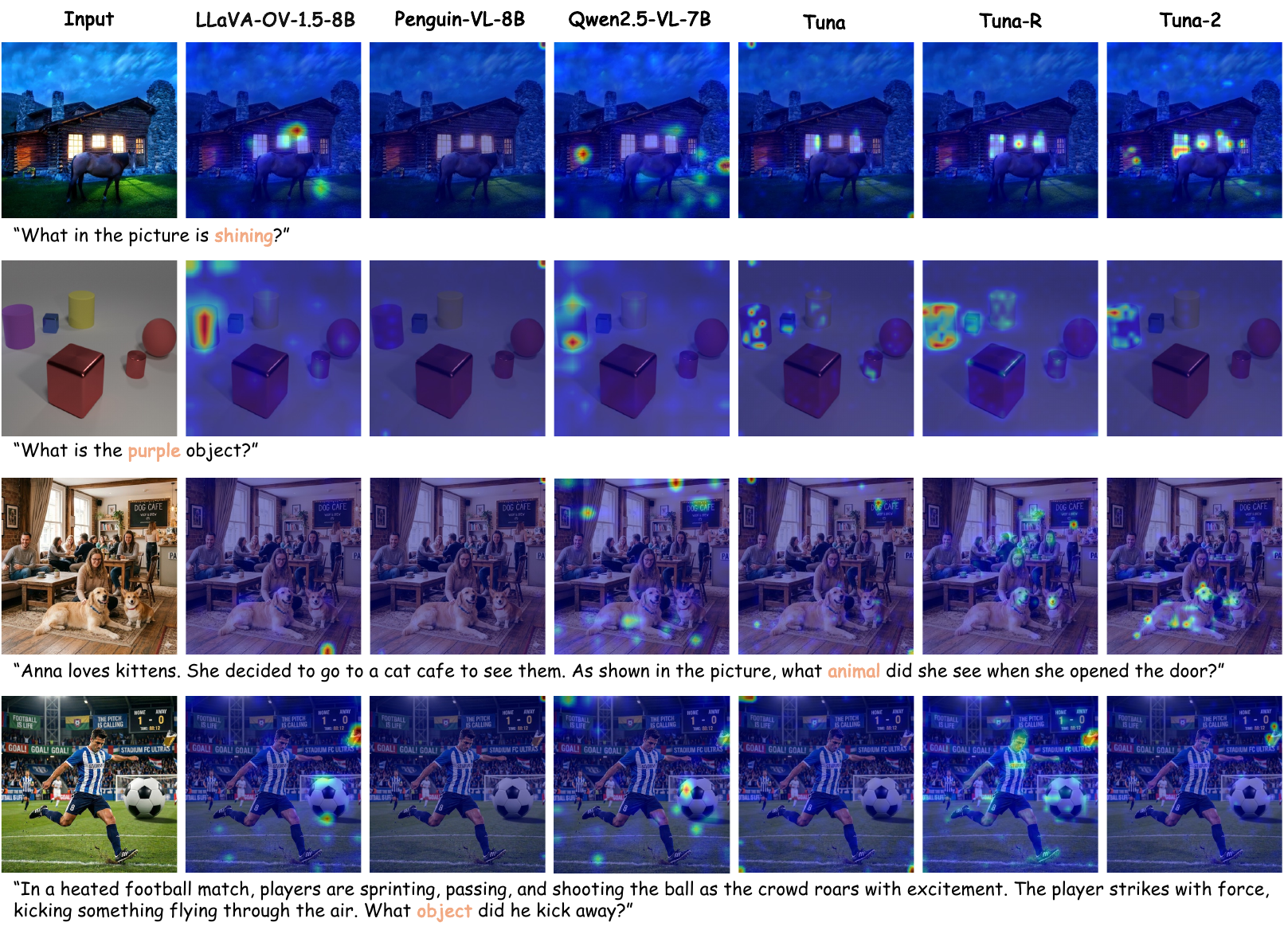}
\vspace{-2.3em}
\caption{Attention map visualization for \modelp, \modeln and other baseline models. \textcolor{red}{Red} area denotes high attention scores and \textcolor{blue}{Blue} area denotes low attention scores.}
\label{fig:attention}
\vspace{-0.5em}
\end{figure*}

\subsection{Analysis: Attention Map Visualizations}
\label{sec:analysis_attention}
To qualitatively analyze and compare the cross-modal alignment capabilities learned by different models, we visualize the attention maps of selected keywords in the input prompt with respect to the input image. We focus on the \tuna model family, including \tuna, \modelp, and \modeln, and compare them with representative encoder-based open-source LMMs, including models with ViT-based encoders, such as LLaVA-OneVision-1.5 \citep{an2025llava} and Qwen2.5-VL \citep{bai2025qwen2}, as well as models with LLM-based encoders, such as Penguin-VL \citep{zhang2026penguin}. The results are shown in Figure~\ref{fig:attention}.

Our first observation is that \modeln exhibits more accurate vision-language alignment in basic perception scenarios. For example, in the ``shining window'' case, \modeln consistently highlights the regions that are semantically associated with ``shining'', while other models tend to provide only coarse or incomplete localization. In the ``purple object'' case, the attention activations of \modeln are well aligned with the target object, whereas other methods often show dispersed attention or spurious activations in irrelevant regions.

We further construct examples with misleading linguistic contexts to evaluate whether models overly rely on language priors. For example, in the ``dog cafe'' case, the prompt suggests a cat cafe scenario, whereas the actual animals in the image are dogs. Some models tend to rely on textual cues within the image (e.g. ``dog cafe'' sign) instead of the actual visual subjects. In contrast, \modeln places its attention on the truly relevant visual regions, demonstrating more robust cross-modal alignment under misleading linguistic contexts.

In the more challenging counterintuitive ``football match'' case, we introduce both strong linguistic priors and salient visual distractors. Specifically, the prompt contains strong football-related cues, such as ``football match'', ``players'', and ``kicking''. The image also includes a large and visually salient football as a distractor. However, the actual object being kicked by the player is a glass cup. Most models are easily misled by either the textual prior or the salient distractor and therefore attend to incorrect regions. In contrast, \modeln accurately localizes the key object that is consistent with the question semantics, showing stronger robustness in such counterintuitive settings.

Overall, as an encoder-free native multimodal architecture, \modeln produces more accurate and stable attention distributions across all cases. These results suggest that \modeln learns more reliable visual representations, leading to more consistent cross-modal alignment and better robustness to misleading language priors and visual distractors.

%% file: tables/table_image_und.tex
\begin{table*}[t]
\centering

\caption{Comparisons between \model and baseline models on multimodal understanding benchmarks. Results with model size greater than 13B are \textcolor{gray}{grayed}. \textbf{Bold}: best results among all UMMs. \underline{Underline}: second-best among all UMMs.}

\vspace{-1em}
\label{tab:img_und}
\footnotesize
\setlength{\tabcolsep}{4pt}
\renewcommand{\arraystretch}{1.05}

\resizebox{\textwidth}{!}{%
\begin{tabular}{@{}l|c|ccccccccc ccc@{}}
\toprule
\multirow{3}{*}{Models} & \multirow{3}{*}{Size} 
  & \multicolumn{9}{c}{General Benchmarks} 
  & \multicolumn{3}{c}{Pixel-centric Benchmarks} \\
\cmidrule(lr){3-11} \cmidrule(lr){12-14}
& & GQA & RealWorldQA & MMVet & MMMU & MMVP & SEED-Bench2\texttt{+} & AI2D & ChartQA & OCRBench 
  & V* & CountBench & VisuLogic \\
\midrule

\multicolumn{14}{c}{\cellcolor[HTML]{EFEFEF}\textit{Understanding-only Models (LMMs)}} \\
LLaVA-1.5 \citep{liu2023visual} & 7B & 62.0 & 54.8 & 32.9 & 35.7 & - & - & 55.5 & 17.8 & 31.8 & - & - & - \\
Qwen-VL-Chat \citep{bai2023qwen} & 7B & 57.5 & 49.3 & 47.3 & 37.0 & - & - & 57.7 & 49.8 & 48.8 & - & - & - \\
LLaVA-OV \citep{li2024llava} & 7B & - & 69.9 & 51.9 & 48.8 & 77.3 & 62.2 & 81.4 & 80.9 & 62.2 & 72.7 & 76.2 & 24.8 \\
Qwen2.5-VL \citep{bai2025qwen2} & 7B & 60.7 & 69.9 & 61.7 & 58.6 & 78.0 & 70.5 & 82.7 & 83.0 & 83.7 & 71.2 & 74.1 & 20.0 \\
\midrule

\multicolumn{14}{c}{\cellcolor[HTML]{EFEFEF}\textit{Composite UMMs}} \\
TokenFlow-XL \citep{qu2025tokenflow} & 14B & 62.5 & 56.6 & - & 43.2 & - & - & - & - & - & - & - & - \\
BLIP3-o \citep{chen2025blip3} & 4B & - & 60.4 & - & 46.6 & - & - & - & - & - & - & - & - \\
Tar \citep{han2025vision} & 7B & 61.3 & - & - & 39.0 & 74.3 & 46.2 & - & - & - & 41.4 & 64.2 & 24.3 \\
X-Omni \citep{geng2025x} & 7B & 62.8 & 62.6 & - & 47.2 & - & - & 76.8 & 81.5 & 70.4 & - & - & - \\
\midrule

\multicolumn{14}{c}{\cellcolor[HTML]{EFEFEF}\textit{Native UMMs}} \\
\textcolor{gray}{BAGEL \citep{deng2025emerging}} & \textcolor{gray}{14B} & \textcolor{gray}{66.4} & \textcolor{gray}{72.8} & \textcolor{gray}{67.2} & \textcolor{gray}{55.3} & \textcolor{gray}{85.0} & \textcolor{gray}{71.9} & \textcolor{gray}{89.2} & \textcolor{gray}{78.5} & \textcolor{gray}{73.3} & \textcolor{gray}{70.2} & \textcolor{gray}{82.5} & \textcolor{gray}{41.7} \\
\textcolor{gray}{Ming-UniVision \citep{huang2025mingunivision}} & \textcolor{gray}{16B} & \textcolor{gray}{59.4} & \textcolor{gray}{59.1} & \textcolor{gray}{64.2} & \textcolor{gray}{40.3} & \textcolor{gray}{71.0} & \textcolor{gray}{56.8} & \textcolor{gray}{82.8} & \textcolor{gray}{76.7} & \textcolor{gray}{72.4} & \textcolor{gray}{48.2} & \textcolor{gray}{76.8} & \textcolor{gray}{26.7} \\
Harmon \citep{wu2025harmonizing} & 1.5B & 58.9 & 49.8 & - & 38.9 & 61.7 & 41.6 & 57.0 & 29.8 & 11.2 & 41.9 & 67.0 & 25.1 \\
JanusFlow \citep{ma2025janusflow} & 1.3B & 60.3 & 41.2 & 36.2 & 29.3 & 67.7 & 39.8 & 54.2 & 42.4 & 53.2 & 42.9 & \underline{78.6} & 22.0 \\
Emu3 \citep{wang2024emu3} & 8B & 60.3 & 57.4 & 23.5 & 31.6 & 71.0 & 44.6 & 70.0 & 69.4 & 68.7 & 53.4 & 65.2 & 24.7 \\
VILA-U \citep{wu2024vila} & 7B & 60.8 & 46.8 & 26.3 & 31.2 & 62.7 & 31.9 & 49.0 & 11.4 & 23.3 & 38.7 & 55.2 & 25.4 \\
Janus-Pro \citep{chen2025janus} & 7B & 62.0 & 58.0 & 41.1 & 41.0 & 73.3 & 56.3 & 71.3 & 25.8 & 59.0 & 47.6 & 53.2 & 23.8 \\
Show-o2 \citep{xie2025show} & 7B & 63.1 & 64.7 & 39.6 & 48.9 & \underline{76.7} & \underline{61.3} & 78.6 & 52.3 & 32.4 & 44.5 & 63.5 & \underline{26.9} \\
OneCat \citep{li2025onecat} & 9B & 63.1 & 65.2 & \textbf{52.2} & 41.9 & 71.3 & \textbf{61.6} & 77.8 & 81.2 & \underline{79.0} & \textbf{63.4} & 34.2 & 24.9 \\
\tuna \citep{liu2025tuna} & 7B & \underline{63.9} & 66.1 & 42.9 & 49.8 & 70.7 & 52.7 & 79.3 & \textbf{85.8} & 74.3 & 52.4 & 73.5 & 22.4 \\
\rowcolor[HTML]{ECF4FF}
\modelp & 7B & 63.5 & \textbf{67.9} & 46.7 & \textbf{51.1} & 74.7 & 58.4 & \underline{79.4} & \underline{85.6} & 78.3 & 57.6 & 77.8 & 26.2 \\
\rowcolor[HTML]{ECF4FF}
\modeln & 7B & \textbf{65.0} & \underline{67.7} & \underline{51.7} & \underline{50.7} & \textbf{77.3} & 61.1 & \textbf{79.6} & \underline{85.6} & \textbf{79.7} & \underline{59.2} & \textbf{81.7} & \textbf{28.8} \\
\bottomrule

\end{tabular}%
}
\vspace{-1em}
\end{table*}

%% file: tables/table_gen.tex
\begin{table}[!t]
\centering

\caption{Image generation results on GenEval and DPG-Bench. ``Col. Attr.'' means ``Color Attribute''. $^\dagger$ refers to methods using LLM rewriters in GenEval. \textbf{Bold}: best results among native UMMs. \underline{Underline}: second-best.}
\label{tab:geneval_dpg}
\vspace{-1em}

\footnotesize
\setlength\tabcolsep{0.7pt}

\resizebox{\textwidth}{!}{%
\begin{tabular}{@{}l|c|ccccccc|cccccc@{}}
\toprule
\multirow{3}{*}{Models} & \multirow{3}{*}{Size} & \multicolumn{7}{c|}{GenEval} & \multicolumn{6}{c}{DPG-Bench} \\
\cmidrule(lr){3-9} \cmidrule(lr){10-15}
 & & 1-Obj. & 2-Obj. & Count & Colors & Position & Col. Attr. & Overall & Global & Entity & Attribute & Relation & Other & Overall \\
\midrule
\rowcolor[HTML]{EFEFEF}
\multicolumn{15}{c}{\cellcolor[HTML]{EFEFEF}\textit{Generation-only Models}} \\
SD3-M \citep{esser2024scaling} & 2B & 0.99 & 0.94 & 0.72 & 0.89 & 0.33 & 0.60 & 0.74 & 87.90 & 91.01 & 89.96 & 80.70 & 88.68 & 84.08 \\
FLUX.1 [dev]$^{\dagger}$ \citep{batifol2025flux} & 12B & 0.98 & 0.93 & 0.75 & 0.93 & 0.68 & 0.65 & 0.82 & 82.10 & 89.50 & 88.70 & 91.10 & 89.40 & 84.00 \\

LongCat-Image \citep{team2025longcat}& 6B & 0.99 & 0.98 & 0.86 & 0.86 & 0.75 & 0.73 & 0.87 & 89.10 & 92.54 & 92.00 & 93.28 & 87.50 & 86.80 \\
Qwen-Image \citep{wu2025qwen}& 20B & 0.99 & 0.92 & 0.89 & 0.88 & 0.76 & 0.77 & 0.87 & 91.32 & 91.56 & 92.02 & 94.31 & 92.73 & 88.32 \\
Seedream 3.0 \citep{gao2025seedream}& - & 0.99 & 0.96 & 0.91 & 0.93 & 0.47 & 0.80 & 0.84 & 94.31 & 92.65 & 91.36 & 92.78 & 88.24 & 88.27 \\

Z-Image-Turbo \citep{cai2025z}&6B & 1.00 & 0.95 & 0.77 & 0.89 & 0.65 & 0.68 & 0.82 & 91.29 & 89.59 & 90.14 & 92.16 & 88.68 & 84.86 \\
\midrule
\rowcolor[HTML]{EFEFEF}
\multicolumn{15}{c}{\cellcolor[HTML]{EFEFEF}\textit{Composite UMMs}} \\
Tar \citep{han2025vision} & 7B & 0.99 & 0.92 & 0.83 & 0.85 & 0.80 & 0.65 & 0.84 & 83.98 & 88.62 & 88.05 & 93.98 & 84.86 & 84.19 \\
BLIP3-o \citep{chen2025blip3} & 8B & - & - & - & - & - & - & 0.84 & - & - & - & - & - & 81.60 \\
UniWorld-V1$^{\dagger}$ \citep{lin2025uniworld} & 12B & 0.98 & 0.93 & 0.81 & 0.89 & 0.74 & 0.71 & 0.84 & 83.64 & 88.39 & 88.44 & 89.27 & 87.22 & 81.38 \\
OmniGen2$^{\dagger}$ \citep{wu2025omnigen2} & 7B & 0.99 & 0.96 & 0.74 & 0.98 & 0.71 & 0.75 & 0.86 & 88.81 & 88.83 & 90.18 & 89.37 & 90.27 & 83.57 \\
\midrule
\rowcolor[HTML]{EFEFEF}
\multicolumn{15}{c}{\cellcolor[HTML]{EFEFEF}\textit{Native UMMs}} \\
MUSE-VL \citep{xie2025muse} & 7B & - & - & - & - & - & - & 0.57 & - & - & - & - & - & - \\
Transfusion \citep{zhou2024transfusion} & 7B & - & - & - & - & - & - & 0.63 & - & - & - & - & - & - \\
Emu3 \citep{wang2024emu3} & 8B & - & - & - & - & - & - & 0.66 & - & - & - & - & - & 81.60 \\
Show-o2 \citep{xie2025show} & 7B & \textbf{1.00} & 0.87 & 0.58 & 0.92 & 0.52 & 0.62 & 0.76 & 89.00 & 91.78 & 89.96 & 91.81 & \textbf{91.64} & 86.14 \\
Janus-Pro \citep{chen2025janus} & 7B & \underline{0.99} & 0.89 & 0.59 & 0.90 & 0.79 & 0.66 & 0.80 & 86.90 & 88.90 & 89.40 & 89.32 & 89.48 & 84.19 \\
HBridge$^{\dagger}$ \citep{wang2025hbridge}& 7B & \textbf{1.00} & \underline{0.96} & 0.80 & \underline{0.94} & 0.77 & 0.78 & 0.87 & \textbf{91.78} & \textbf{91.82} & 90.23 & 90.06 & 88.42 & 85.23 \\
Ming-UniVision \citep{huang2025mingunivision}& 16B & \textbf{1.00} & 0.93 & 0.59 & 0.93 & \textbf{0.92} & 0.70 & 0.85 & - & - & - & - & - & 82.12 \\
BAGEL$^{\dagger}$ \citep{deng2025emerging} & 14B & 0.98 & 0.95 & \textbf{0.84} & \textbf{0.95} & 0.78 & 0.77 & 0.88 & 88.94 & 90.37 & \underline{91.29} & 90.82 & 88.67 & 85.07 \\
Mogao \citep{liao2025mogao} & 7B & \textbf{1.00} & \textbf{0.97} & \underline{0.83} & 0.93 & 0.84 & \underline{0.80} & \underline{0.89} & 82.37 & 90.03 & 88.26 & \underline{93.18} & 85.40 & 84.33 \\
\tuna \citep{liu2025tuna} & 7B & \textbf{1.00} & \textbf{0.97} & 0.81 & 0.91 & \underline{0.88} & \textbf{0.83} & \textbf{0.90} & \underline{90.42} & 91.68 & 90.94 & 91.87 & \underline{90.73} & \textbf{86.76} \\
\rowcolor[HTML]{ECF4FF}
\modelp & 7B & \textbf{1.00} & 0.95 & 0.82 & 0.89 & 0.86 & 0.79 & 0.88 & 86.00 & \underline{91.80} & 91.03 & \textbf{93.48} & 84.89 & 86.35 \\
\rowcolor[HTML]{ECF4FF}
\modeln & 7B  & \underline{0.99} & \underline{0.96} & 0.80 & 0.91 & 0.84 & 0.76 & 0.87 & 89.50 & 91.40 & \textbf{92.07} & 91.91 & 88.81 & \underline{86.54} \\
\bottomrule
\end{tabular}
}

\end{table}

%% file: tables/table_image_third_party.tex
\begin{table*}[t]
\centering

\begin{minipage}[t]{0.40\linewidth}
\centering
\scriptsize

\caption{Evaluation results under GPT-5.4 and Claude Opus 4.7. Quality measures image realism and fine-grained detail richness, while Diversity measures variation among images generated from the same prompt.  \textbf{Bold} denotes the best result and \underline{underlining} denotes the second-best.}
\label{tab:llm_eval}
\vspace{-0.3em}

\setlength{\tabcolsep}{3.5pt}
\renewcommand{\arraystretch}{1.5}
\begin{tabular}{lcccc}
\toprule
& \multicolumn{2}{c}{GPT-5.4} & \multicolumn{2}{c}{Claude Opus 4.7} \\
\cmidrule(lr){2-3} \cmidrule(lr){4-5}
Models & Quality & Diversity & Quality & Diversity \\
\midrule
\textsc{Tuna}\xspace & 22.3\% & 20.6\% & 28.1\% & 28.2\% \\
\modelp              & \textbf{35.7}\% & \underline{30.9}\% & \textbf{37.2}\% & \underline{29.9}\% \\
\modeln              & \underline{32.1}\% & \textbf{48.4}\% & \underline{34.8}\% & \textbf{41.9}\% \\
\bottomrule
\end{tabular}

\end{minipage}
\hfill
\begin{minipage}[t]{0.58\linewidth}
\centering
\scriptsize

\caption{Image editing results on ImgEdit. \textbf{Bold} denotes the best result among unified models and \underline{underlining} denotes the second-best.}
\label{tab:img_edit}

\vspace{-1em}
\setlength{\tabcolsep}{3.7pt}
\renewcommand{\arraystretch}{1.05}
\begin{tabular}{@{}lcccccccccc@{}}
\toprule
Models & Add & Adj. & Ext. & Rep. & Rm. & Bg. & Sty. & Hyb. & Act. & Total \\
\midrule
\rowcolor[HTML]{EFEFEF}
\multicolumn{11}{c}{\textit{Generation-only Models}} \\
FLUX.1 & 4.25 & 4.15 & 2.35 & 4.56 & 3.57 & 4.26 & 4.57 & 3.68 & 4.63 & 4.00 \\
Qwen-Image & 4.38 & 4.16 & 3.43 & 4.66 & 4.14 & 4.38 & 4.81 & 3.82 & 4.69 & 4.27 \\
\midrule
\rowcolor[HTML]{EFEFEF}
\multicolumn{11}{c}{\textit{Unified Models}} \\
OmniGen & 3.47 & 3.04 & 1.71 & 2.94 & 2.43 & 3.21 & 4.19 & 2.24 & 3.38 & 2.96 \\
BAGEL & 3.56 & 3.31 & 1.70 & 3.30 & 2.62 & 3.24 & 4.49 & 2.38 & 4.17 & 3.20 \\
UniWorld & 3.82 & 3.64 & 2.27 & 3.47 & 3.24 & 2.99 & 4.21 & 2.96 & 2.74 & 3.26 \\
OmniGen2 & 3.57 & 3.06 & 1.77 & 3.74 & 3.20 & 3.57 & \underline{4.81} & 2.52 & 4.68 & 3.44 \\
GPT-Image & \textbf{4.61} & \underline{4.33} & \textbf{2.90} & \underline{4.35} & \underline{3.66} & \textbf{4.57} & \textbf{4.93} & \underline{3.96} & \textbf{4.89} & \underline{4.20} \\
\textsc{Tuna}\xspace & \underline{4.43} & \textbf{4.48} & \underline{2.46} & \textbf{4.65} & \textbf{4.55} & \underline{4.52} & 4.69 & \textbf{4.22} & \underline{4.76} & \textbf{4.31} \\
\rowcolor[HTML]{ECF4FF}
\modelp & 4.46 & 4.27 &  2.38& 4.61  &4.48  & 4.44 & 4.54 & 4.06 & 4.43 & 4.18 \\
\rowcolor[HTML]{ECF4FF}
\modeln & 4.34 &  4.13  & 2.22 & 4.53 & 4.42 & 4.36 & 4.58 & 3.91 &4.28  & 4.09 \\
\bottomrule
\end{tabular}

\end{minipage}

\vspace{-0.8em}
\end{table*}

%% file: tables/table_mask.tex
\begin{table}[t!]
\centering
\caption{Ablation study results for \modelp and \modeln, with or without our proposed masking-based feature learning scheme during pretraining.}
\vspace{-1em}
\footnotesize
\setlength{\tabcolsep}{10pt}
\renewcommand{\arraystretch}{1.15}

\label{tab:ablation_mask}
\begin{tabular}{l|cccc}
\toprule
Models & OCRBench  & MMVP & CountBench & GenEval \\
\midrule
\tuna     & 56.9
& 54.0
& 55.6
& 57.2
     \\
\midrule
\modelp w/o Masking      & 58.3 & 56.7 & 57.2 & 55.7 \\
\modelp w/ Masking  & 59.2 & 58.0 & 58.2 & 56.0 \\
\midrule
\modeln w/o Masking         & 55.4 & 52.3 & 53.4 & 47.6 \\
\modeln w/ Masking     & 56.8 & 55.7 & 57.6 & 48.2 \\

\bottomrule
\end{tabular}
\end{table}

%% file: sections/2.related.tex
\section{Related Works}\label{sec:related}
\subsection{Unified Multimodal Models}
Unified multimodal models (UMMs)~\citep{xin2025lumina,you2026llada,tian2026internvl,tong2026beyond,zhang2026nextflow,yang2025mmada,shi2025muddit,shen2025mammothmoda2,xie2025reconstruction,niu2025does,wei2025univideo,wang2025lightfusion, wei2025skywork,cui2025emu3,he2025emma,wang2025ovis,hao2025uni,team2025longcat,ai2025ming,wu2025openuni,fan2025unified} seek to integrate multimodal understanding and generation within a single framework. A prevalent approach for UMMs combines autoregressive (AR) language models for text generation \citep{grattafiori2024llama, bai2023qwen} with diffusion or flow-matching models for visual generation \citep{ho2020denoising, rombach2022high}. While AR models excel at understanding, their integration with high-fidelity generation typically relies on decoupled encoders \citep{chen2025janus, liao2025mogao, deng2025emerging}, leading to representation mismatches and inefficiencies. To address this problem, recent works such as UniTok \citep{ma2025unitok}, TokLip \citep{lin2025toklipmarryvisualtokens}, UniLip \citep{tang2025unilip}, UniFlow \citep{yue2025uniflow}, UAE~\citep{fan2025prism} and OpenVision 3 \citep{zhang2026openvision} pretrain unified vision tokenizers for both semantic understanding and visual reconstruction. Meanwhile, native UMMs like the Show-o series \citep{xie2024show, xie2025show}, \tuna \citep{liu2025tuna}, Ming-UniVision \citep{huang2025ming}, and Transfusion-RAE \citep{tong2026beyond} build unified representations using pretrained VAE \citep{wan2025wan} and representation encoders \citep{zheng2025diffusion}. Despite the significant progress, recent native UMMs still predominantly rely on VAE latents to build their unified representations, which hinders the model's performance on fine-grained visual perception and reasoning. In this work, we systematically investigate pixel-space UMMs with \modelp and \modeln. Concurrent to our work, SenseNova-U1 \citep{diao2026sensenovau1unifyingmultimodalunderstanding} investigates a similar encoder-free architecture (Neo-Unify \citep{diao2026neounify}) with a mixture-of-transformer (MOT) \citep{liang2024mixture} style LLM decoder.

\subsection{Encoder-Free Multimodal Understanding and Generation}
Large multimodal models (LMMs) have traditionally adopted a modular design, combining pixel-space representation encoders with LLM decoders. Early work on LMMs explored integration strategies between representation encoders and LLM decoders, such as using cross-attention in Flamingo \citep{alayrac2022flamingo} and MLP connectors in LLaVA \citep{liu2023visual}. Subsequent research largely followed LLaVA and scaled this paradigm to larger models \citep{wang2024qwen2, bai2025qwen2, wang2025internvl3, an2026video}, datasets \citep{zhang2024video, wiedmann2025finevision, an2025llava}, and diverse understanding tasks \citep{cheng2024videollama,li2024videochat, ren2025vamba}. More recently, an alternative design of employing a monolithic, encoder-free transformer to natively process vision and language has emerged~\citep{wang2025vision,li2026breen,lei2025scalability,shukor2025scaling}. Fuyu \citep{bavishi2023fuyu8b}, EVE \citep{diao2024unveiling}, Chameleon \citep{team2024chameleon}, Mono-InternVL \citep{luo2025mono} and NEO \citep{diao2025pixels} employ simple MLP or patch embedding layers to tokenize raw image pixels into patches, and jointly process these image patches with language tokens using a single transformer. In this work, we show that both the representation encoder-based design and the monolithic design can be integrated into pixel-space UMMs and achieve high performance on multimodal understanding.

Recent visual generation models~\citep{yang2026gloria,chen2024pixart,an2025onestory,an2026vggrpo,zhou2025scaling,qiu2025histream} typically operate in a compressed latent space using KL- or VQ-regularized VAEs \citep{rombach2022high, esser2024scaling, wu2025qwen, esser2021taming, sun2024autoregressive}. Although pixel-space diffusion or flow matching is often considered more challenging, recent work such as PixelFlow \citep{chen2025pixelflow}, PixNerd \citep{wang2025pixnerd}, DiP \citep{chen2025dip}, PixelDiT \citep{yu2025pixeldit} and JiT \citep{li2025back} increasingly suggests that pixel-space flow matching has the potential to match or surpass latent diffusion models. However, these studies are usually limited to small-scale settings (e.g., class-conditioned generation on ImageNet \citep{russakovsky2015imagenet}). We demonstrate in \model that pixel-space flow matching can be scaled up to large-scale unified multimodal pretraining and support free-form text-to-image generation and image editing. 

%% file: sections/5.conclusion.tex
\section{Conclusion}\label{sec:conclu}

We introduced \model, a family of native unified multimodal models that perform multimodal understanding and visual generation directly in pixel space, without relying on VAE encoders or latent diffusion. By combining a unified vision-language backbone with a pixel-space flow matching head, \model supports image understanding, text-to-image generation, and image editing within a single framework. We further instantiated \model with both a representation encoder-based variant and an encoder-free monolithic variant, and showed that both designs achieve strong performance across multimodal understanding and generation benchmarks. Our experiments demonstrate that \model surpasses prior latent-space unified models such as \tuna and Show-o2 on fine-grained visual understanding benchmarks while remaining competitive with state-of-the-art unified models on image generation. Overall, these results highlight the effectiveness and scalability of pixel-space unified multimodal modelling and suggest a promising direction for future native UMMs.